\documentclass[letterpaper]{article} 
\usepackage{aaai2026}  
\usepackage{times}  
\usepackage{helvet}  
\usepackage{courier}  
\usepackage[hyphens]{url}  
\usepackage{graphicx} 
\usepackage{natbib}  
\usepackage{caption} 
\usepackage{algorithm}
\usepackage{algorithmic}

\usepackage{booktabs}
\usepackage{tabularx}
\usepackage{amsmath}
\usepackage{amsfonts}
\usepackage{amssymb}
\usepackage{placeins}
\usepackage{adjustbox}
\usepackage{multirow}
\usepackage{xcolor}
\usepackage{subcaption}
\usepackage{tikz}
\usepackage{enumitem}

\usepackage{newfloat}
\usepackage{listings}
\DeclareCaptionStyle{ruled}{labelfont=normalfont,labelsep=colon,strut=off} 
\floatstyle{ruled}
\newfloat{listing}{tb}{lst}{}
\floatname{listing}{Listing}
\title{Object‑Centric World Models for Causality-Aware Reinforcement Learning}
\author {
    Yosuke Nishimoto\textsuperscript{\rm 1},
    Takashi Matsubara\textsuperscript{\rm 2}
}
\affiliations {
    \textsuperscript{\rm 1}The University of Osaka\\
    \textsuperscript{\rm 2}Hokkaido University\\
    u835711c@ecs.osaka-u.ac.jp, matsubara@ist.hokudai.ac.jp
}

\begin{document}

\maketitle

\begin{abstract}
    World models have been developed to support sample-efficient deep reinforcement learning agents.
    However, it remains challenging for world models to accurately replicate environments that are high-dimensional, non-stationary, and composed of multiple objects with rich interactions since most world models learn holistic representations of all environmental components.
    By contrast, humans perceive the environment by decomposing it into discrete objects, facilitating efficient decision-making.
    Motivated by this insight, we propose \emph{Slot Transformer Imagination with CAusality-aware reinforcement learning} (STICA), a unified framework in which object-centric Transformers serve as the world model and causality-aware policy and value networks.
    STICA represents each observation as a set of object-centric tokens, together with tokens for the agent action and the resulting reward, enabling the world model to predict token-level dynamics and interactions.
    The policy and value networks then estimate token-level cause--effect relations and use them in the attention layers, yielding causality-guided decision-making.
    Experiments on object-rich benchmarks demonstrate that STICA consistently outperforms state-of-the-art agents in both sample efficiency and final performance.
\end{abstract}


\section{Introduction}
\begin{figure*}[t]
    \centering
    \begin{minipage}[b]{0.5\linewidth}
        \centering
        \subfloat[Object-Centric World Model. \label{stica_wm}]{%
            \includegraphics[width=\linewidth,height=10cm,keepaspectratio]{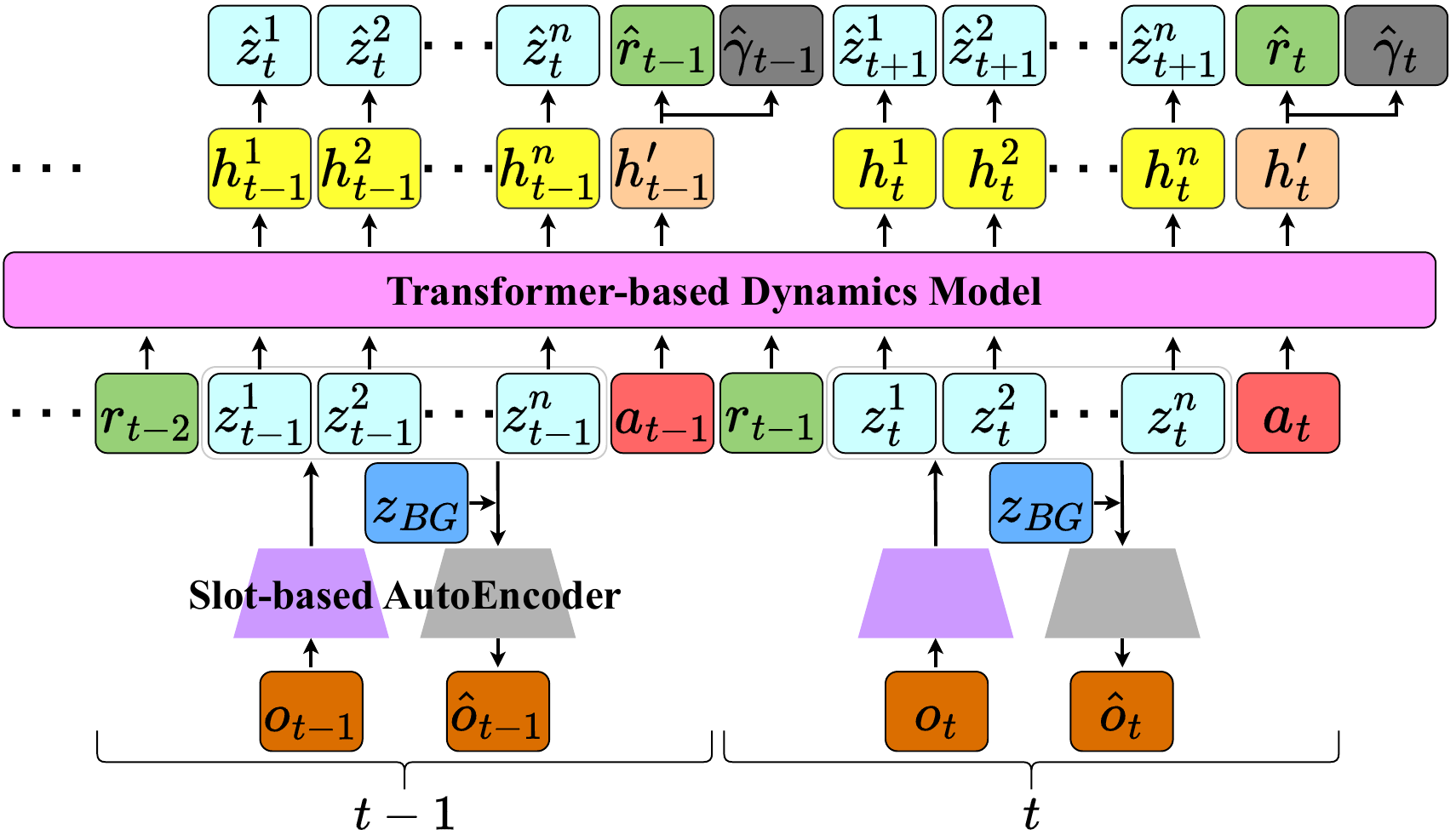}%
        }
    \end{minipage}
    \hfill
    \begin{minipage}[b]{0.49\linewidth}
        \centering
        \subfloat[Object-Centric Representations. \label{slot_image}]{%
            \includegraphics[width=0.9\linewidth,height=4.9cm,keepaspectratio]{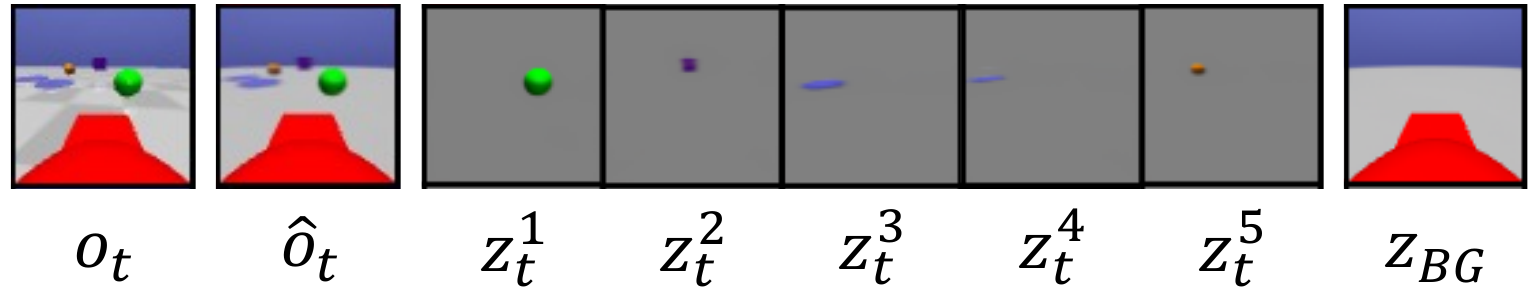}%
        }\\
        \subfloat[Causal Policy and Value Networks. \label{policy}]{%
            \includegraphics[width=0.9\linewidth,height=4.9cm,keepaspectratio]{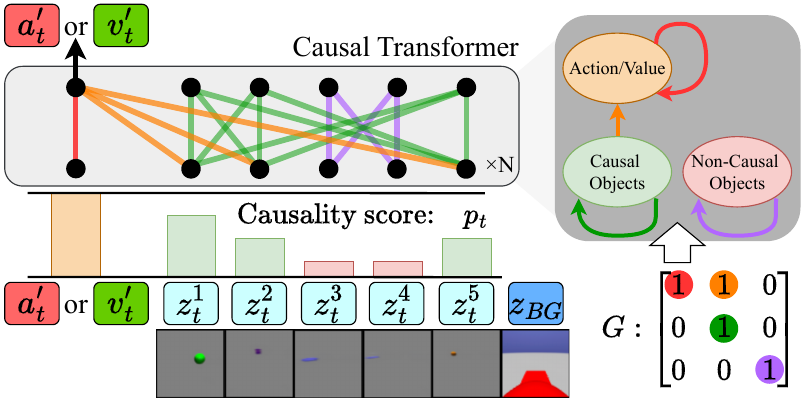}%
        }
    \end{minipage}
    \caption {The architecture and object-centric representations of STICA.
    (a) Object-Centric World Model.
    Slot-based autoencoder extracts object-centric latent states $(z_{t}^1,\dots,z_{t}^n)$ from observation $o_{t}$, excluding static background information $z_{BG}$ at time $t$ ($1\le t \le T)$.
    Transformer-based dynamics model computes hidden states $(h_{1:t}^1, \dots, h_{1:t}^n)$ and $h_{1:t}'$ from latent states $(z_{1:t}^1, \dots, z_{1:t}^n)$, actions $a_{1:t}$, and rewards $r_{1:t-1}$, followed by the multilayer perceptrons (MLPs) that predict the next latent states $(\hat{z}_{t+1}^1, \dots, \hat{z}_{t+1}^n)$, the reward $\hat{r}_t$, and the discount factor $\hat{\gamma}_t$.
    (b) Examples of object-centric representations for Safety Gym benchmark task; the observation $o_t$, its reconstruction $\hat{o}_t$, the reconstructions from the extracted object-centric latent states $(z_t^1,\dots,z_t^5)$, and that from the static background information $z_{BG}$.
    (c) Causal policy and value networks.
    They estimate causal relationships from the latent states to the action or value, based on a causal graph $G$ and causality scores $p_t^k$, and adjust the attention weights within the Transformers accordingly, enabling the causality-aware decision-making.
    The latent states ($z_t^1$, $z_t^2$, and $z_t^5$) of goal-related objects or obstacles are expected to have stronger causal influence on the target token ($a'_t$ or $v'_t$), while latent states ($z_t^3$ and $z_t^4$) of objects irrelevant to task completion have weaker causal influence.
    }
    \label{fig:combined_diagrams}
\end{figure*}
Deep reinforcement learning (RL) has achieved success in a variety of fields~\citep{Mnih:2013ese,Lillicrap2015ContinuousCW,Mnih2016AsynchronousMF,schulman2017ppo,Haarnoja2018SoftAA}, including robotics~\citep{10.5555/2946645.2946684,Andrychowicz_2019,kalashnikov2018qtopt} and autonomous driving~\citep{Sallab_2017,isele2017navigating,Kendall_2019}.
However, achieving high performance requires extensive interactions with the environment, which is costly and inefficient in the real-world tasks when considering real-time operation and physical device failures~\citep{10007800}.

Humans understand and predict real-world dynamics through interaction with the environment~\citep{Sutton:1981a,Friston2010TheFP}.
Inspired by this mechanism, world models were proposed for Model-Based RL (MBRL)~\citep{18,9}.
In this setting, agents train world models to replicate their environments (observations and actions) and optimize their policies within ``imagined'' environments generated by the world models, making their learning sample-efficient.
Earlier MBRL agents have adopted Recurrent Neural Networks (RNNs) as the dynamics model of the world model~\citep{10,11,hafner2025,Kaiser2020Model,deng2022dreamerpro}, whereas recent works have begun to explore the use of Transformers~\citep{Vaswani2017AttentionIA,robine2023twm,micheli2023transformers,10.5555/3692070.3693521,zhang2023storm,burchi2025learning}.
Compared with RNNs, Transformers provide superior learning efficiency, generalization performance, and long-term prediction accuracy.

However, even with Transformers, world models still struggle to replicate environments that are high-dimensional, non-stationary, and composed of multiple objects with their interactions, while such environments are common in real-world applications, including service robots and autonomous driving.
This is because world models learn holistic representations of the environments, which may fail to capture the important relationships and interactions between individual objects~\citep{10.5555/3295222.3295250}.
When placed in such environments, humans perceive the environment by decomposing it into discrete concepts such as objects and events, enabling more efficient and causality-aware decision-making~\citep{SpelkeKinzler2007}.
Incorporating these cognitive mechanisms into world models potentially allow RL agents to operate more effectively even in complex settings.

Motivated by this insight, we propose \emph{Slot Transformer Imagination with CAusality-aware reinforcement learning} (STICA), depicted in Figure~\ref{fig:combined_diagrams}.
This is an RL agent built upon a unified framework in which the world model, policy network, and value network are all implemented using object-centric Transformers, and both the policy and value networks explicitly leverage causal information for more structured and effective decision-making.
The slot-based autoencoder extracts object-centric representations $(z^1_t,\dots,z^n_t)$ from observations $o_t$ while excluding static background information $z_{BG}$, as shown in Figure~\ref{fig:combined_diagrams} (a) and (b).
The Transformer-based dynamics model accepts these representations $(z^1_t,\dots,z^n_t)$, along with agent actions $a_t$ and obtained rewards $r_{t-1}$, as input tokens and predicts token-level dynamics and interactions.
The policy and value networks estimate causal relationships $G$ among the input tokens, thereby facilitating policy learning for causality-aware decision-making, which is aware of ``causality'' in the sense of token-level dependency (not in the context of causal inference) (see Figure \ref{fig:combined_diagrams} (c)).
The main contributions of this paper can be summarized as follows:

\vspace*{0.5mm}\noindent\textbf{High-performance object-centric world model:}
We introduce STICA, a novel framework that significantly outperforms state-of-the-art agents on the Safety-Gym benchmark~\citep{Achiam2019BenchmarkingSE}, a first-person-view, non-stationary, object-rich, 3D environment, as well as on the Object-Centric Visual RL benchmark~\citep{Yoon2023AnII}, a fixed-view, 2D or 3D environment.

\vspace*{0.5mm}\noindent\textbf{Causality-aware decision-making:}
We introduce the policy and value networks that explicitly leverage the token-level dependencies as the information of causal dependencies, enabling decision-making that accurately captures object properties.
By visualizing the attention weights within these networks, we obtain the human-interpretable causal influences that STICA identifies and exploits to guide its actions.

\vspace*{0.5mm}\noindent\textbf{End-to-end learning of object-centric representations:}
By removing static background information from object-centric representation, we enhance the feature-extraction capabilities of autoencoders, which have depended on supervised learning and pretraining in prior studies.
Consequently, this modification enables precise, end-to-end learning of object-centric representations.

\section{Related Work}
\subsection{Model-Based Reinforcement Learning}
\label{mbrl}
Model-based reinforcement learning (MBRL) agents build a world model that replicates the environmental dynamics from experience.
By using this world model for planning, agents improve their generalization capabilities, sample efficiency, and decision-making performance.
Dyna is a classic method that introduced this idea~\citep{18}.
It learns a model of the environment and uses it to refine rewards and value functions.
\citet{9} reintroduced the concept of a world model, employing neural networks to emulate the environment from high-dimensional visual inputs (i.e., images) successfully.

The Dreamer series is a set of MBRL agents that employ the Recurrent State-Space Model (RSSM) as world models, which learns deterministic and stochastic state transitions and allows robust prediction from observations~\citep{20,10,11,hafner2025}.
This series trains world models by reconstructing observations and learns policies using an actor-critic method.
DreamerV2 stabilizes representation learning by adopting categorical latent states and extends to complex, high-dimensional environments with discrete action spaces, such as Atari games~\citep{11}.
DreamerV3 introduces several tricks to further improve the performance across diverse domains~\citep{hafner2025}.
TD-MPC omitted the reconstructions but focused on the prediction of rewards~\citep{pmlr-v162-hansen22a}.
TD-MPC2 redesigned the architecture to stabilize the learning procedure, improving its performance on challenging visual and continuous-control tasks under a fixed hyperparameter set~\citep{Hansen2023TDMPC2SR}.

Several recent works have employed Transformers in place of RSSMs for world models to overcome the long-term memory limitations~\citep{Vaswani2017AttentionIA}.
This approach was pioneered by TransDreamer~\citep{chen2021transdreamer} and advanced by subsequent works: TWM~\citep{robine2023twm}, STORM~\citep{zhang2023storm}, IRIS~\citep{micheli2023transformers}, and $\Delta$-IRIS~\citep{10.5555/3692070.3693521}.
TWISTER~\citep{burchi2025learning} further uses Contrastive Predictive Coding~\citep{DBLP:journals/corr/abs-1807-03748} to promote representation learning suited for long-term prediction.

These MBRL agents learn holistic representations of the environment, which may fail to capture the important relationships and interactions between individual objects~\citep{10.5555/3295222.3295250}.

\subsection{Object-Centric Representation Learning}
\label{ocrl}
Object-centric representation learning is an unsupervised method for decomposing a high-dimensional visual scene into individual objects and learning their individual representations.
Most existing methods employed an autoencoder to extract multiple latent states, referred to as ``slots''.
These slot-based methods have been successfully applied to object detection and tracking in images and videos~\citep{Burgess2019MONetUS,NEURIPS2020_8511df98,singh2022illiterate,seitzer2023bridging,jiang2023objectcentric,wu2023slotdiffusion,Singh2024GuidedLS,kipf2022conditional,elsayed2022savi,singh2022simple,wu2023slotdiffusion}.
This paradigm is also employed for video prediction, which learns the dynamics of individual objects to forecast their future movements and interactions.
For instance, Transformer-based autoregressive models, such as SlotFormer~\citep{wu2023slotformer} and RSM~\citep{nguyen2023reusable}, have demonstrated their robust generalization and impressive long-horizon prediction performance.
SlotSSM~\citep{jiang2024slot} employed a state-space model and showed further performance improvements.

\subsection{Reinforcement Learning with Object-Centric Representation Learning}
Some studies have proposed to combine model-free RL agents with object-centric representation learning~\citep{zadaianchuk2021selfsupervised, Carvalho_2021, yi2022objectcategory, Heravi_2023, Yoon2023AnII, haramati2024entitycentric}.
SMORL~\citep{zadaianchuk2021selfsupervised} and OCARL~\citep{yi2022objectcategory} employ a Transformer to the policy network and operate it on object-centric representations.
OCRL~\citep{Yoon2023AnII} and EIT~\citep{haramati2024entitycentric} further employ a Transformer to the value network.
While this enables decision-making with a focus on individual objects, due to the lack of world model, it does not explicitly leverage their dynamics and interactions.

Object-centric MBRL agents remain less developed, as they require additional resources to learn object-centric representations.
OODP~\citep{Zhu2018ObjectOrientedDP}, COBRA~\citep{watters2019cobra}, OP3~\citep{pmlr-v100-veerapaneni20a}, and SWM~\citep{Singh2021StructuredWB} require numerous episodes generated through random exploration (that is, random action selection without a policy or an estimated value).
FOCUS~\citep{ferraro2023focusobjectcentricworldmodels} and OC-STORM~\citep{zhang2025objectsmatterobjectcentricworld} are based on supervised learning.
SOLD~\citep{mosbach2025soldslotobjectcentriclatent} is pre-trained on external datasets.
Furthermore, most of these methods are not designed to handle non-stationary environments, where new objects appear or existing ones disappear.
This limitation also implies a limited ability to operate in partially observable settings, including first-person views.
Therefore, these agents face significant limitations in terms of generality and sample efficiency.

\section{Method: STICA}
We introduce STICA, an RL agent equipped with an object-centric world model and designed to make more structured and effective decisions by explicitly leveraging causal information about object interactions (see Figure \ref{fig:combined_diagrams}).
STICA is the first MBRL agent that extracts object-centric representations directly from observations without random exploration, supervision, or pretraining.
Its architecture is composed of three components: an object-centric world model, a causal policy network, and a causal value network.
The object-centric world model learns state transitions of latent states of individual objects extracted from visual observations, and simulates the environment to generate imaginary trajectories.
The causal policy and value networks perform on these imaginary trajectories to maximize expected returns.
The training process of STICA follows the standard one that most existing MBRL agents employed.

\subsection{Object-Centric World Model}
\label{world_model}
As illustrated in Figure \ref{fig:combined_diagrams} (a), our object-centric world model is composed of a slot-based autoencoder $\phi$ and a Transformer-based dynamics model $\psi$.
This model is trained on batches of $B$ sequences of $n$ latent states $(z^1_{1:T}, \dots, z^n_{1:T})$, actions $a_{1:T}$, rewards $r_{1:T}$, and discount factors $\gamma_{1:T}$, where the subscripts $1:T$ denote the sequences from $t=1$ to $t=T$.

\paragraph{Slot-based AutoEncoder:}
\label{AutoEncoder}
The slot-based autoencoder $\phi$ is trained to extract object-centric representations and a background representation separately from observations through the reconstruction.
Namely, it is composed of an encoder and a decoder:
\begin{equation}
    \begin{aligned}
      \label{vae}
      &\text{Encoder:} &&p_\phi^z((z^1_t,\dots,z^n_t)|o_t),\\
      &\text{Decoder:} &&p_\phi^{\hat{o}}(\hat{o}_t|(z^1_t,\dots,z^n_t),z_{BG}).
    \end{aligned}
  \end{equation}
Given an observation $o_t$, the encoder obtains $n$ deterministic slots $(s^1_t, \dots, s^n_t)$ using an attention mechanism called Slot Attention~\citep{NEURIPS2020_8511df98}.
Each slot $s^i_t$ serves as 128-dimensional logits to define 16 distinct categorical distributions of 8 categories, from which each latent state $z_t^i$ is sampled as 16 sets of 8-dimensional one-hot vectors~\citep{11}.
An additional learnable time-independent latent state $z_{BG}$ is defined.
This is expected to represent static background information that is independent on time $t$, such as the environment layout and the agent's body.
This trick allows the encoder to extract the object-centric representations of individual objects separately from the static background information, which is crucial for the world model to capture the dynamics of individual objects.
The decoder is implemented as a spatial broadcast decoder~\citep{Watters2019SpatialBD}, which reconstructs the individual RGB images $(\hat{o}^1_t, \dots, \hat{o}^n_t)$ and the individual unnormalized masks $(m^1_t, \dots, m^n_t)$ from the latent states $(z^1_t, \dots, z^n_t)$.
It also reconstructs the background $\hat{o}_{BG}$ from the static background information $z_{BG}$, while the corresponding mask $m_{BG}$ is filled with 0.
A softmax function is applied to the masks $(m^1_t, \dots, m^n_t, m_{BG})$ for obtaining the normalized masks $(M^1_t, \dots, M^n_t, M_{BG})$.

Then, we mix the individual RGB images $(\hat{o}^1_t, \dots, \hat{o}^n_t,$ $\hat{o}_{BG})$ with the normalized masks $(M^1_t, \dots, M^n_t, M_{BG})$ as weights, obtaining a single reconstruction $\hat{o}_t$;
\begin{equation}
    \label{reconstruction}
    \textstyle
      \hat{o}_t = \sum^n_{k=1}{M_t^{k}\odot\hat{o}^{k}_t} + M_{BG}\odot\hat{o}_{BG}.
 \end{equation}
Figure~\ref{fig:combined_diagrams}(b) visualizes examples of the reconstructed individual RGB images $(\hat{o}^1_t, \dots, \hat{o}^n_t, \hat{o}_{BG})$ for latent states $(z^1_t, \dots, z^n_t, z_{BG})$.

\paragraph{Loss for Slot-based AutoEncoder:}
The objective of the slot-based autoencoder $\phi$ is to minimize the following loss function $\mathcal{L}^{ae}_\phi$:
\begin{equation}
    \begin{aligned}
       \label{loss_ae}
       &\textstyle\hspace*{-18mm}\mathcal{L}^{ae}_\phi
       =\mathbb{E}_B[\frac{1}{T}\sum^T_{t=1}\left(\mathcal{J}_{rec.}^t
       +\alpha_1\mathcal{J}_{ent.}^t
       +\alpha_2\mathcal{J}_{cross}^t\right)], \\
      \text{where}\quad
      \mathcal{J}_{rec.}^t &\textstyle=-\text{ln}p_\phi^{\hat{o}}(o_t|(z^1_t,\dots,z^n_t), z_{BG}),\\
      \mathcal{J}_{ent.}^t &\textstyle=-\sum^n_{k=1}{H(p_\phi^z(z^k_t|o_t))},\quad \\
      \mathcal{J}_{cross}^t&\textstyle=\sum^n_{k=1}{H(p_\phi^z(z^k_t|o_t),p_\psi^{\hat{z}}(\hat{z}^k_t|h^k_{t-1}))},
    \end{aligned}
 \end{equation}
with hyperparameters $\alpha_1,\alpha_2\geq0$.
The reconstruction error $\mathcal{J}_{rec.}^t$ ensures that the latent states $(z^1_t,\dots,z^n_t)$ and $z_{BG}$ contain sufficient information to reconstruct the observation $o_t$.
The entropy term $\mathcal{J}_{ent.}^t$ is a regularization term that prevents the distribution of each latent state $z^k_t$ from being deterministic.
The cross-entropy term $\mathcal{J}_{cross}^t$ is another regularization term that aligns each extracted latent state $z^k_t$ with the latent state $\hat{z}^k_t$ predicted by the Transformer-based dynamics model $\psi$.
Here, as the slot encoder extracts the latent states $(z^1_t,\dots,z^n_t)$ in a random order, the cross-entropy term $\mathcal{J}_{\text{cross}}^t$ is computed after rearranging the indices $1,\dots,n$ to minimize the L1 distance between the logits of the categorical distributions for $z^k_t$ and $\hat{z}^k_t$.

\begin{figure*}[htbp]
  \centering
  \begin{minipage}{0.72\textwidth}
    \centering
    \foreach \x in {pointgoal1, pointgoal2, pointbutton1, pointbutton2, cargoal1, cargoal2, carbutton1, carbutton2} {
      \begin{minipage}{0.1099\textwidth}
        \centering
        \includegraphics[width=\linewidth]{image/\x.pdf}\\
        {\centering}
      \end{minipage}
    }
    \vspace{0.2em}
    \foreach \x/\lab in {
    pointgoal1_fixed/PointGoal1,
    pointgoal2_fixed/PointGoal2,
    pointbutton1_fixed/PointButton1,
    pointbutton2_fixed/PointButton2,
    cargoal1_fixed/CarGoal1,
    cargoal2_fixed/CarGoal2,
    carbutton1_fixed/CarButton1,
    carbutton2_fixed/CarButton2} {
    \begin{minipage}{0.1099\textwidth}
    \centering
    \includegraphics[width=\linewidth]{image/\x.pdf}\\
    {\scriptsize \centering \lab}
    \end{minipage}
    }
  \end{minipage}
  \hfill
\begin{minipage}{0.26\textwidth}
  {\scriptsize
    \begin{minipage}{0.3\textwidth}
      \includegraphics[width=\linewidth]{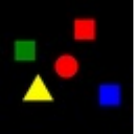}\\
      \centering Obj.~Goal.
    \end{minipage}
    \hspace{0.3em}
    \begin{minipage}{0.3\textwidth}
      \includegraphics[width=\linewidth]{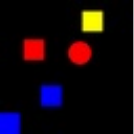}\\
      \centering Obj.~Inter.
    \end{minipage}
    \hspace{0.3em}
    \begin{minipage}{0.3\textwidth}
      \includegraphics[width=\linewidth]{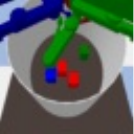}\\
      \centering Obj.~Reach.
    \end{minipage}
  }
\end{minipage}
  \caption{(Left) Eight tasks from the Safety Gym benchmark. Top panels show first-person views for training. Bottom panels show fixed-view images for reference and not for training.
  (Right) Three tasks from the OCVRL benchmark.}
  \label{benchmark_img}
\end{figure*}

\paragraph{Transformer-based Dynamics Model:}
The Transformer-based dynamics model $\psi$ predicts the future latent states $(z^1_{t+1}, \dots, z^n_{t+1})$, the reward $r_{t}$, and the discount factor $\gamma_{t}$ from the past reward $r_{s-1}$, latent states $(z^1_s, \dots, z^n_s)$, and action $a_s$ for $s=t,t-1,t-2,\dots$.
This model $\psi$ is composed of the following modules:
\begin{equation}
    \begin{aligned}
    \label{11}
    &\text{Aggregation model:}\\
    &\quad (h^1_t,\dots,h^n_t, h'_t) = f_\psi^h(r_{1:t-1},(z^1_{1:t},\dots,z^n_{1:t}),a_{1:t}),\hspace*{-8cm}\\
    &\text{Latent state predictor:}&& p_\psi^{\hat{z}}(\hat{z}^k_{t+1}|h^k_t) \quad \text{for } k = 1, \dots, n,\\
    &\text{Reward predictor:}&& p_\psi^{\hat{r}}(\hat{r}_t|h'_t),\\
    &\text{Discount predictor:}&& p_\psi^{\hat{\gamma}}(\hat{\gamma}_t|h'_t),
    \end{aligned}
\end{equation}
where the aggregation model $f_\psi^h$ is implemented using Transformer-XL~\citep{dai-etal-2019-transformer}, and the others $p_\psi$ using multilayer perceptrons (MLPs).

The aggregation model $f_\psi^h$ feeds the history of the reward $r_{1:t-1}$, latent states $(z^1_{1:t}, \dots, z^n_{1:t})$ and action $a_{1:t}$ as tokens and predicts the hidden states $(h^1_t, \dots, h^n_t)$ and $h'_t$ as the outputs of the current latent states $(z^1_t, \dots, z^n_t)$ and action $a_t$ in a deterministic manner.
In this predictor, causal masking is applied to ensure self-attention layers not to access future time steps.
The positional encoding depends only on time $t$ and not on the indices $1,\dots,n$ of the latent states to ensure the equivariance to the order of the latent states.

The latent state predictor $p_\psi^{\hat{z}}(\hat{z}^k_{t+1} \mid h^k_t)$ and the reward predictor $p_\psi^{\hat{r}}(\hat{r}_t|h'_t)$ output a categorical distribution and a normal distribution through the reparameterization trick, respectively~\citep{19}.
The discount predictor $p_\psi^{\hat{\gamma}}(\hat{\gamma}_t \mid h'_t)$ outputs a value in the range of $[0, 1]$.

\paragraph{Loss for Transformer-based Dynamics Model:}
The objective of the Transformer-based dynamics model $\psi$ learns the transitions of the latent states $(z^1_t, \dots, z^n_t)$ by minimizing the following loss function:
\begin{equation}
    \begin{aligned}
       \label{17}
      &\textstyle\hspace*{-21mm}\mathcal{L}^{dyn}_\psi
       =\mathbb{E}_B[\frac{1}{T}\sum^T_{t=1}\left(\mathcal{J}_{cross}^{t+1}
       +\beta_1\mathcal{J}_{rew.}^t
       +\beta_2\mathcal{J}_{dis.}^t\right)],\\
      \text{where}\quad
      \mathcal{J}_{cross}^{t+1}&\textstyle=\sum^n_{k=1}{H(p_\phi^z(z^k_{t+1}|o_{t+1}),p_\psi^{\hat{z}}(\hat{z}^k_{t+1}|h^k_{t}))},\\
      \mathcal{J}_{rew.}^t     &\textstyle=-\text{ln}(p_\psi^{\hat{r}}(r_t|h'_t)),\\
      \mathcal{J}_{dis.}^t     &\textstyle=-\text{ln}(p_\psi^{\hat{\gamma}}(\gamma_t|h'_t)),
    \end{aligned}
 \end{equation}
with hyperparameters $\beta_1,\beta_2\geq0$, and $\gamma_t=0$ for episode ends and $\gamma_t=\gamma$ otherwise.
The cross-entropy term $\mathcal{J}_{cross}^{t+1}$ ensures that each predicted latent state $\hat{z}^k_{t+1}$ is close to the extracted one $z^k_{t+1}$.
This term $\mathcal{J}_{cross}^{t+1}$ is also computed after rearranging the indices of the latent states.
The negative log-likelihoods $\mathcal{J}_{rew.}^t$ and $\mathcal{J}_{dis.}^t$ optimizes the predictions of the reward $\hat{r}_t$ and discount factor $\hat{\gamma}_t$, respectively, where the discount factor $\hat{\gamma}_t$ is considered as a Bernoulli variable.

\subsection{Policy and Value Networks with Causal Attention}
\label{policy_value}
\paragraph{Transformer-based Policy and Value Networks:}
STICA learns the policy on latent states $(\hat{z}^1_t, \dots, \hat{z}^n_t)$ generated by our object-centric world model using an advantage actor-critic (A2C) method~\citep{Mnih2016AsynchronousMF}.
The advantage function is estimated using the generalized advantage estimator (GAE)~\citep{Schulmanetal_ICLR2016}.
The policy network $\pi_\theta(a'_t|(z^1_t,\dots,z^n_t))$ and value network $V_{\xi}(v'_t|(z^1_t,\dots,z^n_t))$ are implemented using two separate Transformers (see Figure \ref{fig:combined_diagrams} (c)).
Each Transformer takes as input tokens the latent states $(z^1_t, \dots, z^n_t)$ and an extra learnable token, whose corresponding output is used for the estimate of the target (action $a'_t$ or value $v'_t$).
All latent states $(z^1_t, \dots, z^n_t)$ share the same positional encoding, ensuring the Transformer's output to be independent of their order (in other words, making the output equivariant to the order of the latent states).

\paragraph{Causal Attention:}
An environment contains both causal objects, which are relevant for action selection and value estimation, and non-causal objects, which are not.
Selecting an action requires attending to reward-relevant objects like goals and obstacles, while paying limited attention to others, such as the floor.
Similarly, value estimation can ignore non-causal objects entirely.
Attention mechanisms within the policy and value networks can leverage these causal relationships.
We explicitly represent these causal relationships using a causal graph matrix $G$ as follows:
\begin{equation}
    \label{causal_mask}
   G =
   \begin{bmatrix}
      1 & 1 & 0 \\
      0 & 1 & 0 \\
      0 & 0 & 1
   \end{bmatrix}
      \begin{array}{l}
        \text{\!\!\!\!\} policy or value} \\
        \text{\!\!\!\!\} causal object} \\
        \text{\!\!\!\!\} non-causal object} \\
   \end{array}
\end{equation}
Each element $G_{i,j}$ indicates a causal relationship from object $j$ to object $i$, where the index takes $1$ for the target (policy or value), $2$ for a causal object, and $3$ for a non-causal object.
For example, $G_{1,2}=1$ indicates the existence of a causal relationship from the causal objects to the target.

To leverage these causal relationships for the policy and value networks, we estimate the causality score $p^k_t\in[0,1]$ for each latent state $z^k_t$ at time step $t$ by using an MLP, as shown in Figure~\ref{fig:combined_diagrams} (c).
The causality score $p^k_t$ is a probability that the latent state $z^k_t$ represents a causal object.
Based on these estimates, we adjust the attention weights in the Transformers.
Specifically, we define the weight matrix $W_t$ as follows:
\begin{equation}
   \label{mask_weight}
   W_{t} =
   \begin{bmatrix}
   1 & 0 & 0 \\
   0 & p^1_{t} & 1 - p^1_{t} \\
   \vdots & \vdots & \vdots \\
   0 & p^n_{t} & 1 - p^n_{t}
   \end{bmatrix}
   \begin{array}{lll}
   &\hspace{-6mm}\} &\hspace*{-7mm}\text{policy or value}\\
   \rule{0pt}{1.1em}&\multirow{3}{*}{\hspace{-6mm}\scalebox{1}[3.6]{\}}} \\
   \rule{0pt}{1.1em}&&\hspace*{-7mm}\text{latent states}\\
   \rule{0pt}{1.1em}&\\
   \end{array},
\end{equation}
where the first row corresponds to the target token (policy or value), and the remaining rows correspond to the tokens for the latent states $(z^1_t, \dots, z^n_t)$.
Then, the matrix $W_tGW_t^\top$ represents the causal relationships among the tokens, where the $(i,j)$ element of $W_t$ represents the influence of the $j$-th token on the $i$-th token.
Given this, we define the causal attention $\text{CA}_t$ as follows:
\begin{equation}
   \begin{aligned}
   \label{pruning_attention}
   \text{CA}_t
   = \text{Norm} \left(\text{softmax} \left(\frac{Q_tK^\top_t}{\sqrt{d}}\right) \odot W_tGW_t^\top\right) V_t.
   \end{aligned}
\end{equation}
Here, $Q_t$, $K_t$, and $V_t$ denote the queries, keys, and values at time step $t$, respectively, and $d$ is the dimensionality of the key vectors.
$\odot$ denotes element-wise multiplication, and $\text{Norm}(\cdot)$ denotes row-wise L1 normalization of the input matrix.
The matrix $W_t G W_t^\top$ scales the original attention weights $\text{softmax} \left( \frac{Q_t K_t^\top}{\sqrt{d}} \right)$ to reflect the causal relationships.
Each attention layer in the Transformers of the policy and value networks uses this causal attention $\text{CA}_t$ instead of the standard attention mechanism.

\begin{figure*}[t]
    \centering
    \includegraphics[width=0.84\linewidth]{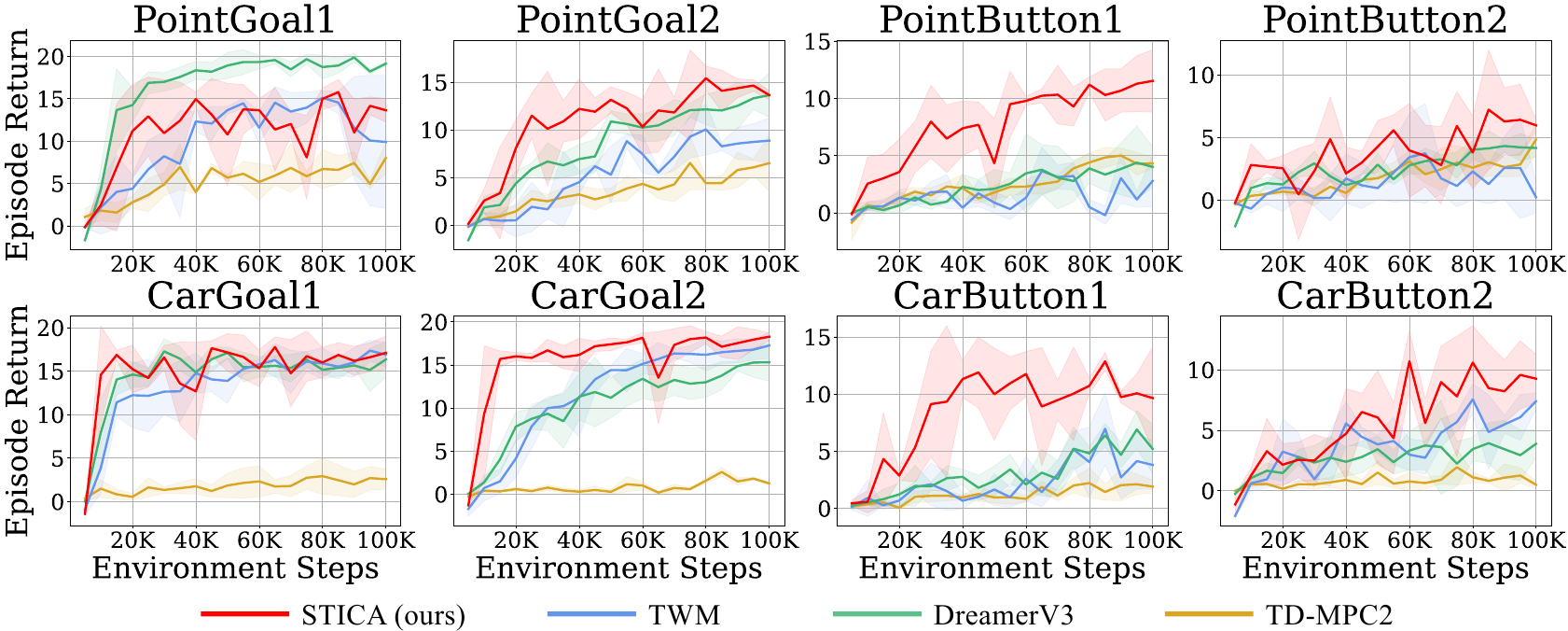}
    \caption{Training curves for Safety Gym benchmark.}
    \label{safety_gym_graph}
\end{figure*}

\begin{table*}[t]
    \centering
    \begin{adjustbox}{width=0.70\textwidth,center}
    \begin{tabular}{lrrrrrrrrrr}
    \toprule
    \multirow{2}{*}{\textbf{Method}} &
    \multicolumn{4}{c}{\textbf{Point}} &
    \multicolumn{4}{c}{\textbf{Car}} &
    \multirow{2}{*}{\textbf{Mean}} &
    \multirow{2}{*}{\textbf{\shortstack{Normalized\\Mean}}} \\
    \cmidrule(lr){2-5} \cmidrule(lr){6-9}
    &
    \multicolumn{1}{c}{{\textbf{Goal1}}} &
    \multicolumn{1}{c}{{\textbf{Goal2}}} &
    \multicolumn{1}{c}{{\textbf{Button1}}} &
    \multicolumn{1}{c}{{\textbf{Button2}}} &
    \multicolumn{1}{c}{{\textbf{Goal1}}} &
    \multicolumn{1}{c}{{\textbf{Goal2}}} &
    \multicolumn{1}{c}{{\textbf{Button1}}} &
    \multicolumn{1}{c}{{\textbf{Button2}}} &
    & \\
    \midrule
    PPO           & 5.00                             & 4.43                            & \underline{4.94}               & 3.78                         & 3.15                          & 1.46                        & 0.95                          & 1.51                        & 3.15                         & 1.00 \\
    TWM           & 9.89                             & 8.87              & 2.79                           & 0.23                         & \underline{16.89}             & \underline{17.25}           & 3.79                          & \underline{7.40}            & 8.39                         & 3.83 \\
    DreamerV3     & \textbf{19.13}       & \textbf{13.64}      & 4.01                           & 4.16                         & 16.32                         & 15.30                       & \underline{5.20}              & 3.87                        & \underline{10.20}            & \underline{4.06} \\
    TD-MPC2       & 8.00                             & 6.50                            & 4.31                           & \underline{4.84}             & 2.57                          & 1.24                        & 1.90                          & 0.48                        & 3.73                         & 1.15 \\
    \midrule
    STICA(ours)   & \underline{13.63}                & \textbf{13.64}      & \textbf{11.52}     & \textbf{5.97}    & \textbf{17.09}    & \textbf{18.27}  & \textbf{9.65}     & \textbf{9.25}   & \textbf{12.38}   & \textbf{5.49} \\
    \bottomrule
    \end{tabular}
    \end{adjustbox}
    \caption{Agent Scores on Safety Gym benchmark.}\label{safety_gym_table}
\end{table*}

\section{Experiments and Results}
\subsection{Benchmarks and Baselines}
\paragraph{Safety Gym Benchmark:}
For validating the performance of STICA, we used eight object-rich 3D tasks from the Safety Gym benchmark~\citep{Achiam2019BenchmarkingSE}, as summarized in Figure~\ref{benchmark_img} (left).
The agent receives first-person images as observations $o_t$, exemplified in the top panels.
For an overview of the environment layout, please refer to the fixed-view images in the bottom panels, while these are not used for experiments.

The tasks are classified into two types: Goal and Button.
The Goal task is to reach a target (green cylinder) area among many static obstacles.
After reaching the target, its location is randomly reset, while the rest of the layout is preserved.
The Button task is to press a target (green) circular button among many other (yellow) circular buttons and dynamic cubic obstacles.
Once the target button is pressed, another button is selected as a new target, with all other objects remaining fixed.
The agent receives rewards for moving towards the current targets densely and for completing the tasks sparsely.
We have two types of agents: Point and Car.
The Point agent has separate actuators for rotation and forward/backward movement, while the Car agent is equipped with two independently-driven parallel wheels and a free-rolling rear wheel.
Each task has two difficulty levels: Level 1 (less objects) and Level 2 (more objects).
We measured the average undiscounted reward return, $\hat{J}(\pi) = \frac{1}{E} \sum_{i=1}^{E} \sum_{t=0}^{T_{\text{ep}}} r_t$, over $E=10$ episodes of $T_{\text{ep}}=1000$ steps, as described in \citet{Achiam2019BenchmarkingSE}.

While the Safety Gym benchmark was originally designed to evaluate the safetiness of RL agents via both rewards and costs (i.e., failures), our goal is to develop an object-centric world model and an RL agent capable of causality-aware decision-making.
Consequently, we use only the average undiscounted return as our performance metric.
Nevertheless, the Safety Gym benchmark provides a suitable evaluation setting because its environments feature the key challenges: high-dimensional, partially observable first-person views; non-stationary targets; and rich interactions with many dynamic obstacles.

See Appendices A.1 for more details about benchmarks and B for the experimental settings.

\paragraph{Object-Centric Visual RL Benchmark:}
We also evaluated STICA on two 2D and one 3D tasks from the Object-Centric Visual RL (OCVRL) benchmark~\citep{Yoon2023AnII}.
The agent receives fixed-view images as observations $o_t$, as exemplified in Figure~\ref{benchmark_img} (right).
Object Goal Task imposes the agent (red circle) to move towards a target object (blue square).
If the agent touches other objects, it receives no reward, and the episode is terminated.
Object Interaction Task requires the agent (red circle) to push a specific object (blue square) to the fixed goal position (blue square at the bottom left corner).
In these tasks, the actions are to move the agent in four discrete directions.
Object Reaching task is a 3D version of Object Goal Task, where the agent has a three-fingered robot arm and is tasked to touch the target object with one finger.
The action adjusts the positions of three joints of the finger.
In any tasks, the agent receives a sparse reward when it reaches the target object or pushes the target object to the goal position.

Although these environments are fully observable and stationary (i.e., no objects appear or disappear), we use this benchmark to demonstrate the generality of STICA.

\paragraph{Baselines:}
We also evaluated three MBRL agents as baselines.
TWM is a Transformer-based world model that learns holistic state transitions~\citep{robine2023twm}.
DreamerV3~\citep{hafner2025} and TD-MPC2~\citep{Hansen2023TDMPC2SR} are state-of-the-art MBRL agents that also learn holistic representations of the environment.
Because TD-MPC2 is specifically designed for tasks with continuous action spaces, we omitted it from the OCVRL benchmark's 2D tasks.
See Appendix A.2 for more details about baselines.
\subsection{Results}
\paragraph{Safety Gym Benchmark:}
We summarized the time courses of the reward returns for the Safety Gym benchmark in Figure \ref{safety_gym_graph}.
Solid lines denote the averages over three runs, while shaded areas denote the standard deviations.
STICA achieved the best performance in all tasks except for PointGoal1 in terms of both the final performance and the learning speed.
See also the final average returns in Table \ref{safety_gym_table}, where bold and underlined numbers indicate the best and the second best performances, respectively.
Normalized mean is the average after being normalized against the performance by Proximal Policy Optimization (PPO), as suggested by the benchmark~\citep{Achiam2019BenchmarkingSE}.
We attribute STICA's superior performance to the object-centric and causality-aware modules as verified later.
PointGoal1 is a simpler task with fewer and less diverse objects than others, where holistic representation may be sufficient, and STICA did not show a significant advantage.

Please refer to Appendix C for other visualizations.

\begin{figure}[t]
    \centering
    \includegraphics[width=0.9\linewidth]{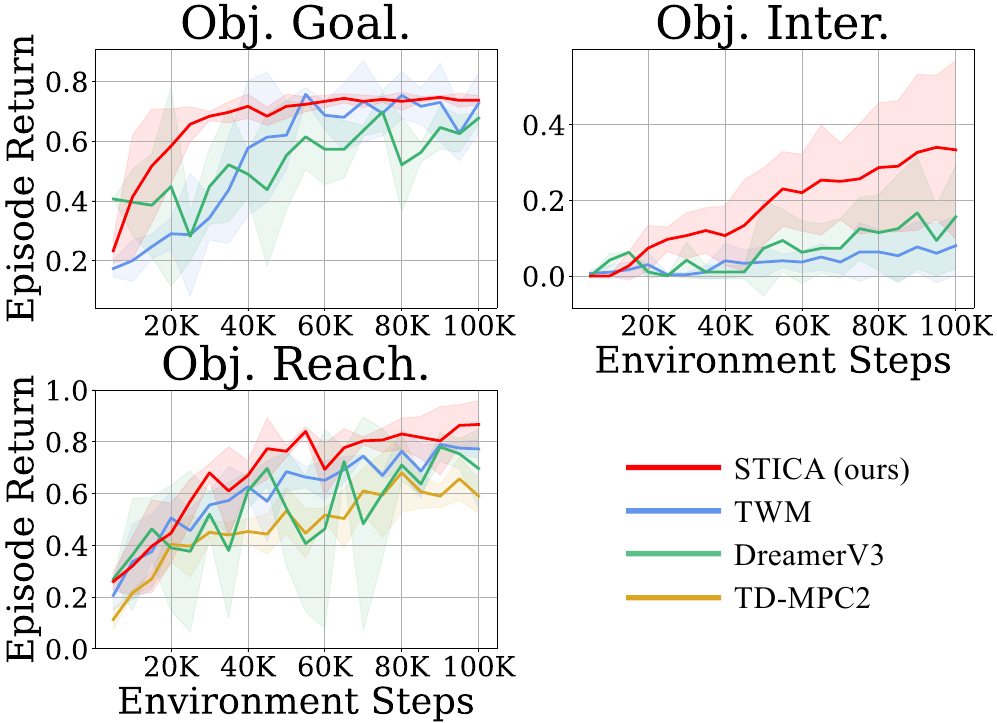}
    \caption{Training curves for the OCVRL benchmark.}
    \label{ocrl_graph}
\end{figure}

\paragraph{Object-Centric Visual RL Benchmark:}
Figure~\ref{ocrl_graph} shows the training curves for the OCVRL benchmark, and Table~\ref{ocrl} summarizes the final success rates averaged over three runs.
STICA achieves superior performance on all tasks.
In particular, STICA achieved a drastically better result on Object Interaction Task, where the agent directly interacts with objects.
In such case, the object-centric world model and the object-centric causal value network act synergistically.

\begin{table}[t]
    \centering
    \begin{adjustbox}{width=0.34\textwidth,center}
    \begin{tabular}{lrrr}
    \toprule
    \textbf{Method} &
    \multicolumn{1}{c}{{\textbf{Obj. Goal.}}} &
    \multicolumn{1}{c}{{\textbf{Obj. Inter.}}} &
    \multicolumn{1}{c}{{\textbf{Obj. Reach.}}} \\
    \midrule
    TWM           & \underline{0.727}                 & 0.080               & \underline{0.772}\\
    DreamerV3     & 0.677                             & \underline{0.156}   & 0.697\\
    TD-MPC2       & -                                 & -                   & 0.590\\
    \midrule
    STICA(ours)   & \textbf{0.737}                    & \textbf{0.333}   & \textbf{0.867}\\
    \bottomrule
    \end{tabular}
    \end{adjustbox}
    \caption{Final success rates on the OCVRL benchmark.}
    \label{ocrl}
\end{table}

\begin{figure}[t]
    \centering
    \includegraphics[width=0.96\linewidth]{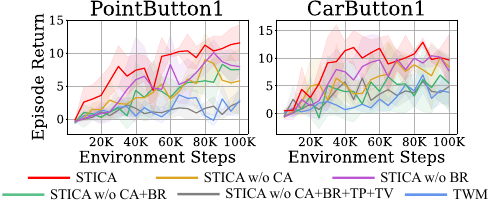}
    \caption{Ablation studies on the PointButton1 and CarButton1 from the Safety Gym benchmark.}
    \label{ablation}
\end{figure}

\subsection{Ablation Study}
To assess the contributions of the modules in STICA, we conducted an ablation study on two tasks (PointButton1 and CarButton1) from the Safety Gym benchmark, and summarize the results in Figure \ref{ablation}.
We refer to the use of the static background information $z_{BG}$ as the background removal, as it removes the background information from the object-centric representations.
We consider the following variants: STICA w/o CA, which omits the causal attention mechanism; STICA w/o BR, which omits the background removal; STICA w/o CA+BR, which omits both; and STICA w/o CA+BR+TP+TV, which further replaces the Transformer-based policy and value networks with their original architectures.
A variant that omits object-centric representations effectively corresponds to TWM.

We observe that adopting an object-centric world model alone yields only a marginal performance improvement (STICA w/o CA+BR+TP+TV vs.\ TWM).
The object-centric Transformers for the policy and value networks produce a noticeable gain in performance (STICA w/o CA+BR vs.\ STICA w/o CA+BR+TP+TV).
The background removal leads to a modest improvement (STICA vs.\ STICA w/o BR, or STICA w/o CA vs.\ STICA w/o CA+BR).
The introduction of causal attention yields a substantial performance increase (STICA vs.\ STICA w/o CA).

\begin{figure}[t]
    \centering
    \includegraphics[width=0.76\linewidth]{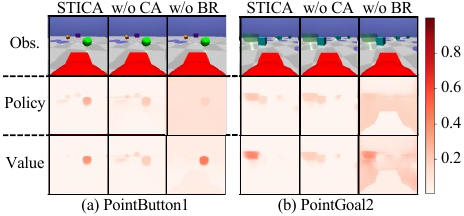}
    \caption{Visualization of the attention weights in the policy and value networks.}
    \label{policy_vis}
\end{figure}

\subsection{Visualization}
We also examined the contributions of causal attention.
From the policy and value networks, we obtained the average attention weight from the token for each latent state $z^k_t$ to the target token (action $a'_t$ or value $v'_t$) in the attention layers, denoted by $A^k_t$.
For visibility, we visualized the average attention weights $A^k_t$ in the regions of the corresponding normalized masks $M^k_t$, that is, $\sum^n_{k=1}{M^k_t A^k_t}$, in Figure \ref{policy_vis}.

We observed that STICA allocates strong and selective attention to task-relevant (that is, causal) objects in both networks, compared to STICA w/o CA or w/o BR.
In particular, the value network in STICA attends exclusively to the reward-yielding target objects (i.e., the green button or green cylinder area), with no appreciable attention paid to other objects.
The policy network also focuses on the target object, but it pays some attention to the other objects, which is necessary for action selection.
In contrast, the policy and value networks in STICA w/o CA fail to focus on the target object, and instead, they pay much attention to various objects in the environment.
Moreover, the policy and value networks in STICA w/o BR also pay much attention to the background, which is completely irrelevant to the task.
Overall, STICA is able to allocate selective attention to objects related to tasks by using causal attention and background removal.

\section{Conclusion}
In this paper, we introduced STICA, a novel RL agent with an object-centric world model and causality-aware policy and value networks.
By decomposing observations into individual objects while removing static background information, STICA learns object-centric representations and their individual dynamics from scratch.
Its causal attention enables the policy and value networks to focus on task-relevant objects, leading to more structured and effective decision-making.
Through these modules, STICA significantly outperforms state-of-the-art RL agents on object-rich benchmarks in both sample efficiency and final performance.

\section*{Acknowledgments}
This study was partly supported by JST PRESTO (JPMJPR24TB, JPMJPR21C7), CREST (JPMJCR1914), ASPIRE (JPMJAP2329), Moonshot R\&D (JPMJMS2033-14), and JSPS KAKENHI (24K15105), Japan, and achieved through the use of large-scale computer systems at the D3Center, Osaka University.


\onecolumn
\appendix
\setcounter{secnumdepth}{2}
\setcounter{section}{0}
\renewcommand\thesection{\Alph{section}}
\renewcommand\thesubsection{\Alph{section}.\arabic{subsection}}
\renewcommand\labelenumi{\thesubsection}

\section{Details on Benchmarks and Baselines}

\subsection{Benchmarks}\label{benchmarks}
\subsubsection{Safety Gym benchmark}
Despite the rapid development of RL agents, they still struggle to replicate environments that are high-dimensional, non-stationary, and composed of multiple objects with their interactions, while such environments are common in real-world tasks, such as service robots and autonomous driving.
We believe that the key to solving this problem is to learn object-centric representations for world models, which learns the dynamics of individual objects and their interactions, rather than holistic representations that learn the dynamics of the entire environment.
While several studies have already developed object-centric world models, their main focus was on fixed-view (fully observable) environments and did not take into account the appearance of new objects or the disappearance of existing ones~\citep{Zhu2018ObjectOrientedDP,watters2019cobra,pmlr-v100-veerapaneni20a,Singh2021StructuredWB,ferraro2023focusobjectcentricworldmodels,zhang2025objectsmatterobjectcentricworld,mosbach2025soldslotobjectcentriclatent}.
Also, these models depend on random exploration, supervision, or pretraining, meaning that the object-centric representations are not learned autonomously.
To examine these aspects, we employed the Safety Gym benchmark~\citep{Achiam2019BenchmarkingSE}.

The Safety Gym benchmark provides multiple tasks for continuous control in 3D environments, where an agent is tasked with navigating toward a goal while avoiding obstacles.
The observations are, by default, first-person-view RGB images of size $64 \times 64$.
This benchmark includes three types of tasks (Goal, Button, and Push) with three types of agents (Point, Car, and Doggo) at two difficulty levels (Level 1 and Level 2).
We conduct experiments on eight tasks: PointGoal1, PointGoal2, PointButton1, PointButton2, CarGoal1, CarGoal2, CarButton1, and CarButton2, where the task names are composed of the agent type, task type, and difficulty level.
Completing the Push tasks or controlling the Doggo agent is substantially more difficult and requires 10 to 100 times more environmental steps than other tasks, as reported in \citet{as2022constrained}, and prior works often excluded some of these tasks (e.g., ~\citet{huang2024safedreamer}); we also excluded these tasks.

The environments for the Goal tasks contain the following types of objects.
A green cylinder is a target object.
A light blue box is a moving obstacle which avoids the agent from reaching the target, and a blue circle is a non-physical object.
All objects are randomly set at the beginning of each episode, and the agent is rewarded for reaching the target object.
After reaching the target object, its location is randomly reset, while the rest of the layout remains unchanged.

The environments for the Button tasks contain the following types of objects.
A green ball is a target button.
A yellow ball is a non-target button, and a purple cube is a moving obstacle.
Both objects avoid the agent from reaching the target button.
A blue circle is a non-physical object.
All objects are randomly set at the beginning of each episode, and the agent is rewarded for reaching the target button.
After reaching the target button, another yellow button is randomly selected as a new target, while the rest of the layout remains unchanged.

\paragraph{Object-Centric Visual RL benchmark}
The Object-Centric Visual RL (OCVRL) benchmark~\citep{Yoon2023AnII} provides four 2D tasks (Object Goal, Object Interaction, Object Comparison, and Property Comparison) and one 3D task (Object Reaching).

In the 2D tasks, the observations are fixed-view RGB images of size $64 \times 64$.
The agent is always represented as a red circle.
At the beginning of each episode, the agent's position is fixed at the center of the environment.
The action set consists of four actions: move up, move down, move left, and move right.

In the Object Goal Task, shapes and colors of objects are chosen from [square, triangle, star] and [blue, green, yellow, red], respectively.
The target object is always a blue square, and only one target object is present in the environment at a time.
The other objects are randomly generated.
The agent is tasked to move towards the target object.
If the agent touches other objects, it receives no reward, and the episode is terminated.
In the Object Interaction Task, colors of objects are chosen from [blue, green, yellow, red] and their shapes are always square.
The target object and a goal position are always represented by blue squares.
The other objects are randomly generated.
The distance of each object from the walls is at least the size of the object, allowing the agent to push the target in any direction.
The agent is tasked to push the target object to the goal position.
The episode is terminated when the task is completed or the episode length exceeds a given limit.

In the Object Comparison Task and Property Comparison Task, the target object differs from episode to episode.
The Object Comparison Task imposes the agent to move towards a single object that differs in color or shape.
The Property Comparison Task is similar to the Object Comparison Task but more difficult; the non-target objects always share either color or shape with the target object.
These tasks are not aligned with the motivation behind our proposed method, causal attention, and are therefore excluded from our experiments.

In the 3D Object Reaching Task, the environment consists of a tri-finger robotic arm and multiple cubes with different colors and sizes.
The color of cubes are chosen from [blue, green, yellow, red].
The agent's task is to reach and manipulate the target object, which is always blue.
The observations are rendered RGB images of size $64 \times 64$.
The actions are limited to the manipulation of the third finger of the robotic arm, with the other two fingers fixed in the upright position.

\clearpage

\subsection{Baselines}\label{baseline}
We employed three MBRL agents as baselines for comparison with our proposed MBRL agent, STICA.
All baselines are capable of learning the dynamics of high-dimensional visual scenes.

\paragraph{Transformer-based World Model}
Transformer-based world model (TWM)~\citep{robine2023twm} is an MBRL agent that serves as the backbone of our proposed STICA.
Unlike conventional recurrent neural network-based world models, as its name implies, TWM employs the Transformer-XL architecture as its world model~\citep{dai-etal-2019-transformer}.
Specifically, this Transformer functions as an autoregressive model of latent states extracted by an autoencoder, taken actions, and experienced or predicted rewards.
Our proposed STICA extends TWM by incorporating object-centric representations and causal attention.

\paragraph{DreamerV3}
DreamerV3~\citep{hafner2025} is one of the state-of-the-art MBRL agents.
The learning efficiency and performance of DreamerV3 increase monotonically with model size, ranging from 12 million to 400 million parameters.
Because STICA has around 35 million parameters, we used the 50-million-parameter variant of DreamerV3 in our experiments.
For further implementation details and insights, we refer the reader to~\citet{hafner2025}.

\paragraph{TD-MPC2}
TD-MPC2~\citep{Hansen2023TDMPC2SR} is also one of the state-of-the-art MBRL agents.
Unlike DreamerV3, TD-MPC2 omits the reconstruction of high-dimensional visual inputs but focuses on the prediction of rewards.
This method employs temporal difference (TD) learning to predict future returns in the latent space and uses model predictive control (MPC) to optimize action sequences.
For our experiments, we used the default 5-million-parameter variant as recommended by the authors for single-task RL problems.
Since TD-MPC2 is specifically designed for tasks with continuous action spaces, we do not evaluate it on the 2D tasks (i.e., Object Goal and Object Interaction in the OCVRL benchmark).

\newpage

\section{Experimental Details}
\label{ExperimentalDetails}
STICA takes approximately five days to train with one NVIDIA A100 GPU.
We list the hyperparameters used in our experiments in Table \ref{hyperparameters} and the numbers of parameters in STICA in Table \ref{model_size}.
\begin{table}[h]
\centering
\caption{Hyperparameters of STICA.}
\label{hyperparameters}
\scriptsize
\begin{tabular}{lcr}
\toprule
\footnotesize{Name} & \footnotesize{Symbol} & \footnotesize{Value} \\
\midrule

\multicolumn{3}{c}{\footnotesize{Slot-based AutoEcoder}} \\
\midrule
{\footnotesize Number of Slots} & {\footnotesize $n$} & {\footnotesize $5$} \\
{\footnotesize Slot Size} & {} & {\footnotesize $128$} \\
{\footnotesize Number of Slot Attention iterations} & {} & {\footnotesize $4$} \\
\midrule
\multicolumn{3}{c}{\footnotesize{Transformer-based Dynamics Model}} \\
\midrule
{\footnotesize Transformer Embedding Size} & {\footnotesize  } & {\footnotesize $256$} \\
{\footnotesize Number of Transformer Layers} & {\footnotesize  } & {\footnotesize $10$} \\
{\footnotesize Number of Transformer Heads} & {\footnotesize } & {\footnotesize $4$} \\
{\footnotesize Transformer Feedforward Size} & {\footnotesize } & {\footnotesize $1024$} \\
{\footnotesize Number of Latent State Predictor Units} & {\footnotesize } & {\footnotesize $4\times512$} \\
{\footnotesize Number of Reward Predictor Units} & {\footnotesize  } & {\footnotesize $4\times256$} \\
{\footnotesize Number of Discount Predictor Units} & {\footnotesize  } & {\footnotesize $4\times256$} \\
\midrule
\multicolumn{3}{c}{\footnotesize{Causal Policy and Value Networks}} \\
\midrule
{\footnotesize Causal Transformer Embedding Size} & {\footnotesize  } & {\footnotesize $256$} \\
{\footnotesize Number of Causal Transformer Layers} & {\footnotesize  } & {\footnotesize $10$} \\
{\footnotesize Number of Causal Transformer Heads} & {\footnotesize } & {\footnotesize $4$} \\
{\footnotesize Causal Transformer Feedforward Size} & {\footnotesize } & {\footnotesize $512$} \\
{\footnotesize Causal Relationship Predictor Units} & {\footnotesize } & {\footnotesize $512, 128, 64$} \\
\midrule
\multicolumn{3}{c}{\footnotesize{World Model Learning}} \\
\midrule
{\footnotesize World Model Batch Size} & {\footnotesize $B$} & {\footnotesize $64$} \\
{\footnotesize History Length} & {\footnotesize $T$} & {\footnotesize $16$} \\
{\footnotesize Discount Factor} & {\footnotesize $\gamma$} & {\footnotesize $0.99$} \\
{\footnotesize Coefficient of $\mathcal{J}_{ent.}$} & {\footnotesize  $\alpha_1$} & {\footnotesize $5.0$} \\
{\footnotesize Coefficient of $\mathcal{J}_{cross}$} & {\footnotesize  $\alpha_2$} & {\footnotesize $0.03$} \\
{\footnotesize Coefficient of $\mathcal{J}_{rew.}$} & {\footnotesize  $\beta_1$} & {\footnotesize $10.0$} \\
{\footnotesize Coefficient of $\mathcal{J}_{dis.}$} & {\footnotesize  $\beta_2$} & {\footnotesize $50.0$} \\
{\footnotesize Learning Rate of World Model} & {\footnotesize  } & {\footnotesize $0.0001$} \\
\midrule
\multicolumn{3}{c}{\footnotesize{Policy and Value Networks Learning}} \\
\midrule
{\footnotesize Imagination Batch Size} & {} & {\footnotesize $400$} \\
{\footnotesize Imagination Horizon} & {} & {\footnotesize $15$} \\
{\footnotesize Policy Network Learning Rate} & {\footnotesize  } & {\footnotesize $0.0001$} \\
{\footnotesize Critic Network Learning Rate} & {\footnotesize  } & {\footnotesize $0.00001$} \\
{\footnotesize GAE Parameter} & {} & {\footnotesize $0.95$} \\
\midrule
\multicolumn{3}{c}{\footnotesize{Interaction with Environments}} \\
\midrule
{\footnotesize Environment Steps} & {} & {\footnotesize $100$K} \\
{\footnotesize Observation Image Size} & {} & {\footnotesize $3\times64\times64$} \\
{\footnotesize Action Repeats (Safety Gym / OCVRL)} & {} & {\footnotesize $2$ / $1$} \\
{\footnotesize Max Episode Length (Safety Gym / OCVRL)} & {} & {\footnotesize $1000$ / $100$} \\
\bottomrule
\end{tabular}
\end{table}
\begin{table}[H]
    \centering
    \caption{Number of parameters in STICA.}
    \label{model_size}
    \scriptsize
    \begin{tabular}{lcr}
        \toprule
        {\footnotesize Name} & {\footnotesize Symbol} & {\footnotesize Parameters}\\
        \midrule
        {\footnotesize Slot-based AutoEncoder} & {\footnotesize $\phi$} & {\footnotesize 6.5M} \\
        {\footnotesize Transformer-based Dynamics Model} & {\footnotesize $\psi$} & {\footnotesize 15M} \\
        {\footnotesize Policy Network} & {\footnotesize $\theta$} & {\footnotesize 6.5M} \\
        {\footnotesize Value Network} & {\footnotesize $\xi$} & {\footnotesize 6.5M} \\
        \midrule
        {\footnotesize Total} & {\footnotesize } & {\footnotesize 35M} \\
    \bottomrule
    \end{tabular}
\end{table}
\clearpage
\section{Additional Visualizations}
\label{imagination}
In this section, we visualize example trajectories generated by STICA on the benchmarks in Figure~\ref{imagination_image}.
In all cases, each slot tracks a single object and predicts its motion.
In Figure~\ref{imagination_image} (d), the agent first turns right and then turns left again, as evidenced by the motion of the yellow ball in the row of $\hat z_t^5$.
In the row of $\hat z_t^4$, the purple cube exits the frame and subsequently reappears.
This demonstrates that the Transformer-based world model retains and leverages past states precisely for prediction.
We set the number of slots to five.
When fewer than five objects appear in the frame (as in Figure~\ref{imagination_image} (c)), the slot-based autoencoder extracts an irrelevant object (the horizon in this case) instead, and the slot effectively remains unused.
The same applies to Figure~\ref{imagination_image} (j).
Such slots are ignored during the decision-making stage via causal attention.
In Figure~\ref{imagination_image} (c), in the row of $\hat z_t^5$, the green button turns yellow upon contact with the agent.
This indicates that the task has been completed and the environment has been updated, yet tracking remains successful.
In this way, STICA can adapt to non-stationary environments.
In Figure~\ref{imagination_image} (g), in the row of $\hat z_t^4$ and $\hat z_t^5$, after the purple cube disappears, the blue circle is shown in a different slot; because slots are unordered, this reassignment does not pose any problem.

\begin{figure}[htbp]
  \centering
  \begin{subfigure}[t]{0.48\linewidth}
    \centering
    \includegraphics[width=\linewidth]{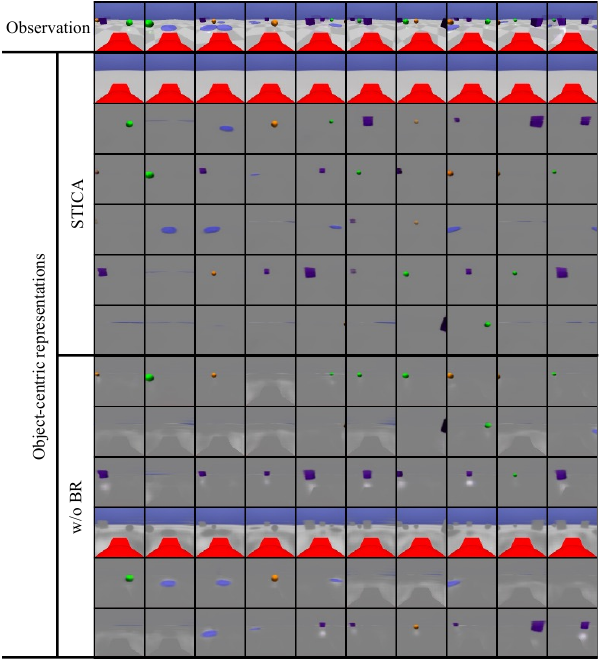}
    \caption{PointButton1}
    \label{fig:pointbutton1_slot}
  \end{subfigure}
  \hfill
  \begin{subfigure}[t]{0.48\linewidth}
    \centering
    \includegraphics[width=\linewidth]{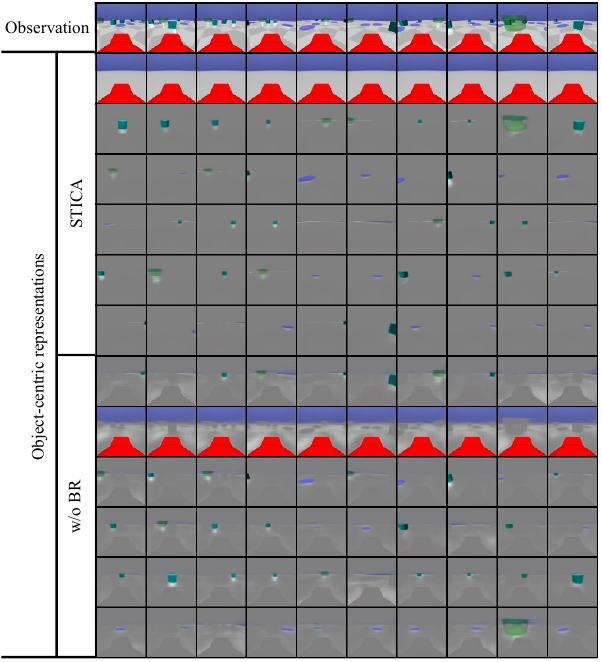}
    \caption{PointGoal2}
    \label{fig:pointgoal2_slot}
  \end{subfigure}
  \caption{Visualization of object-centric representations for random observations.}
  \label{fig:slot}
\end{figure}

Figure~\ref{fig:slot} shows examples of object-centric representations for random observations.
STICA consistently succeeds in assigning individual objects to separate slots.
Recall that ``w/o BR'' denotes STICA without the latent state dedicated for the static background information.
Even under this condition, STICA w/o BR allocates a specific slot to the background (more precisely, to the background and the agent's body); however, as suggested by residual shadows of other objects, the separation is not perfect.
Conversely, the background can be seen faintly in slots for other objects.
Thus, we can conclude that, by using a dedicated latent space for the background, STICA effectively prevents the objects and background from mixing with each other.

\begin{figure}[htbp]
  \centering
  \begin{subfigure}[t]{0.48\linewidth}
    \centering
    \includegraphics[width=\linewidth]{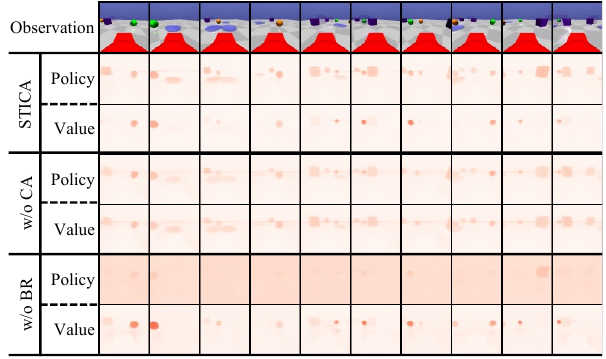}
    \caption{PointButton1}
    \label{fig:pointbutton1_attn}
  \end{subfigure}
  \hfill
  \begin{subfigure}[t]{0.48\linewidth}
    \centering
    \includegraphics[width=\linewidth]{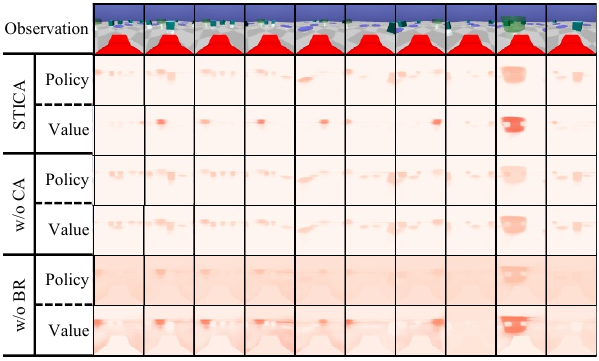}
    \caption{PointGoal2}
    \label{fig:pointgoal2_attn}
  \end{subfigure}
  \caption{Visualization of attention weights in the policy and value networks for random observations.}
  \label{fig:attn_maps}
\end{figure}

Figure~\ref{fig:attn_maps} illustrates the attention weights for random observations in both the policy and value networks.
Each column corresponds to the same column in Figure~\ref{fig:slot}.
In the leftmost column of (a), only three objects are present, so STICA assigns two slots to meaningless objects (i.e., the horizon), as shown in Figure~\ref{fig:slot}.
The policy network of STICA nonetheless focuses exclusively on the three objects, ignoring the horizon, whereas the policy network of STICA w/o CA also attends to the horizon.
Similar patterns are observed in the other columns, and the attention weights of STICA w/o CA are often blurred.
The difference is even more pronounced for STICA w/o BR: although it pays some attention to objects, it attends broadly across the entire scene, including the background.
While the impact of the omission of BR appears limited in Figure~\ref{fig:slot}, the mixture of the objects and background has a substantial adverse effect on the attention maps of both the policy and value networks.
These observations suggest that because the background contains some object information, the networks accidentally exploit it, yet, since it does not carry sufficient task-relevant information, this misuse leads to inappropriate decision-making.

\newpage
\begin{figure}[htbp]
  \centering
  \begin{minipage}[t]{0.48\textwidth}
    \centering
    \includegraphics[width=\textwidth]{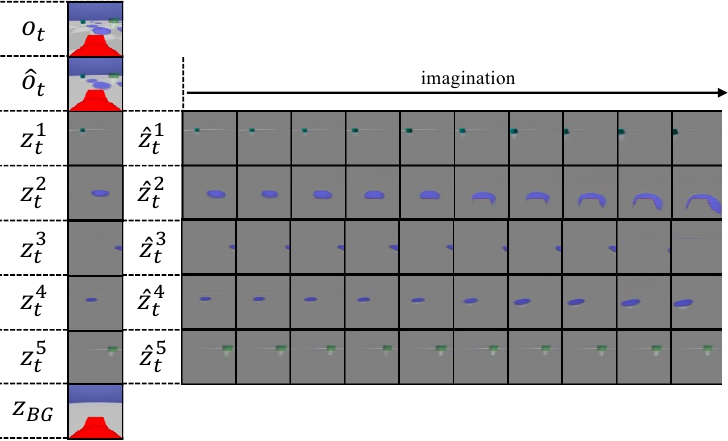}
    \vspace{-8mm}
    \caption*{(a) PointGoal1}
    \vspace{5mm}
  \end{minipage}\hfill
  \begin{minipage}[t]{0.48\textwidth}
    \centering
    \includegraphics[width=\textwidth]{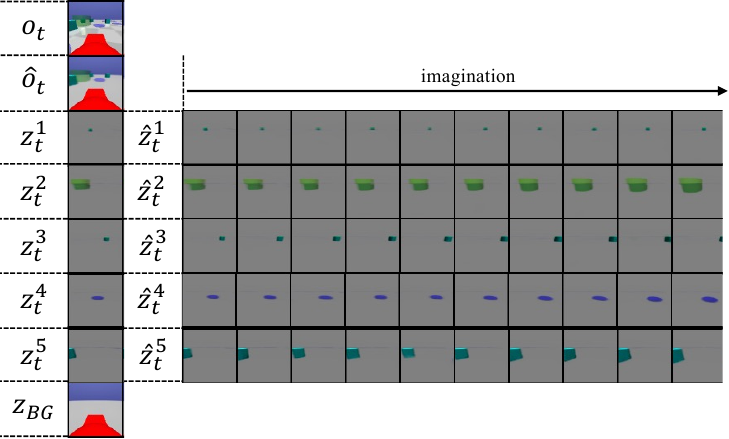}
    \vspace{-8mm}
    \caption*{(b) PointGoal2}
    \vspace{5mm}
    \vspace{-8mm}
  \end{minipage}
  \begin{minipage}[t]{0.48\textwidth}
    \centering
    \includegraphics[width=\textwidth]{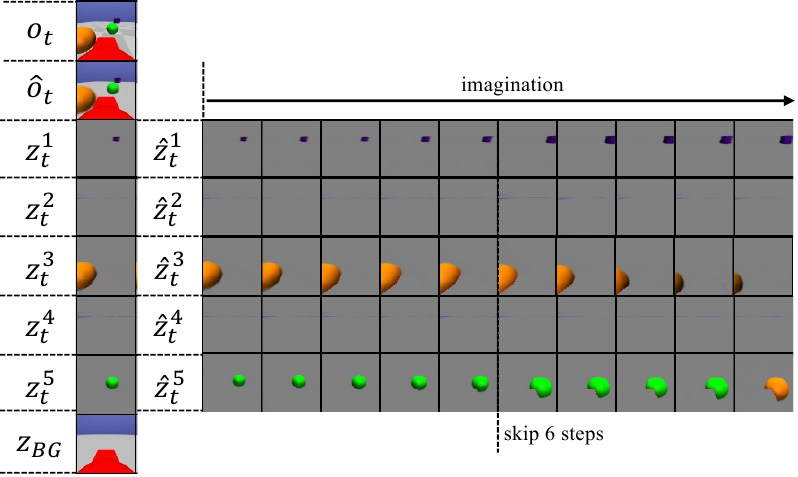}
    \vspace{-8mm}
    \caption*{(c) PointButton1}
    \vspace{5mm}
  \end{minipage}\hfill
  \begin{minipage}[t]{0.48\textwidth}
    \centering
    \includegraphics[width=\textwidth]{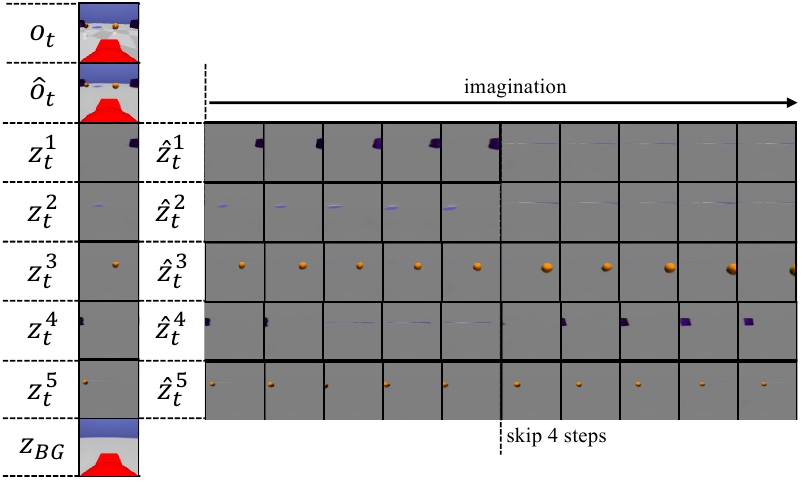}
    \vspace{-8mm}
    \caption*{(d) PointButton2}
    \vspace{5mm}
  \end{minipage}
  \begin{minipage}[t]{0.48\textwidth}
    \centering
    \includegraphics[width=\textwidth]{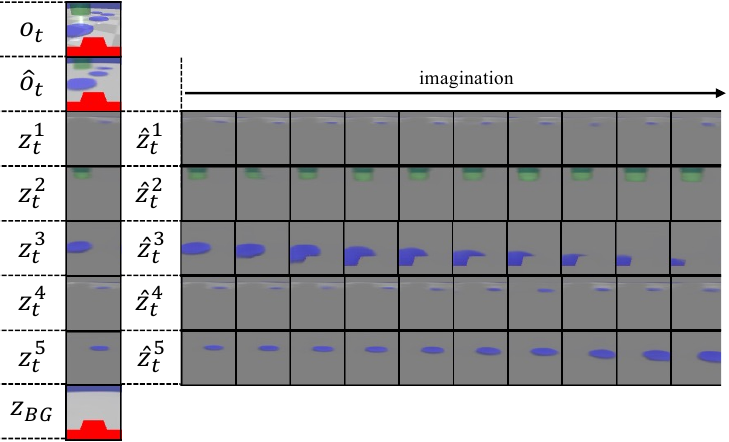}
    \vspace{-8mm}
    \caption*{(e) CarGoal1}
  \end{minipage}\hfill
  \begin{minipage}[t]{0.48\textwidth}
    \centering
    \includegraphics[width=\textwidth]{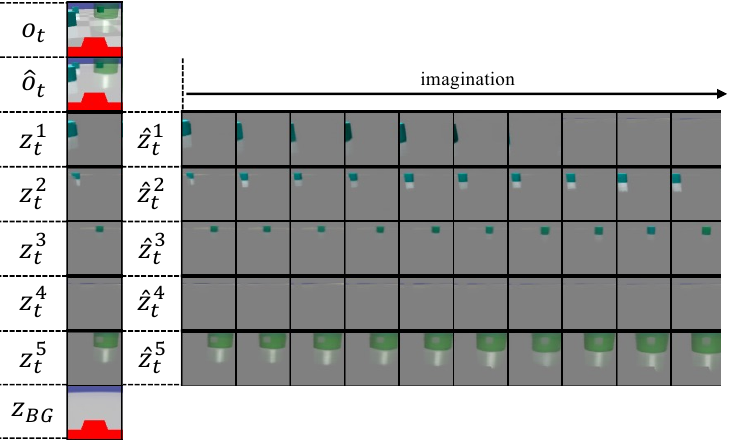}
    \vspace{-8mm}
    \caption*{(f) CarGoal2}
  \end{minipage}
  \caption{Examples of generated trajectories by the object-centric world model $\psi$ of STICA.
  (a)--(f) The Safety Gym benchmark (Continued on next page). In (c) and (d), some time steps are omitted for visibility.}\label{imagination_image}
\end{figure}
\begin{figure}[h]
  \ContinuedFloat
  \centering
  \begin{minipage}[t]{0.48\textwidth}
    \centering
    \includegraphics[width=\textwidth]{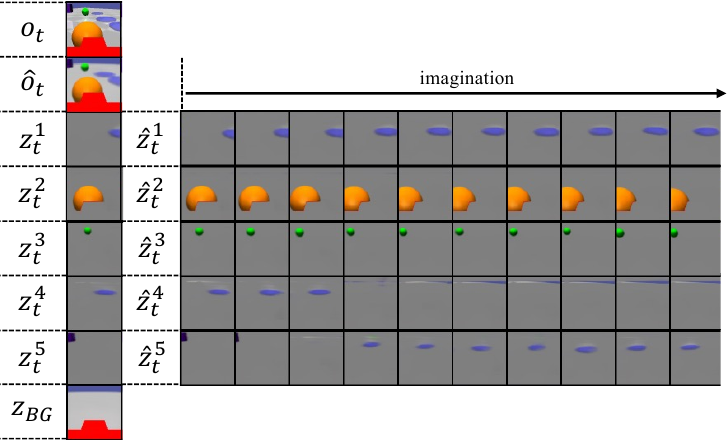}
    \vspace{-8mm}
    \caption*{(g) CarButton1}
    \vspace{5mm}
  \end{minipage}\hfill
  \begin{minipage}[t]{0.48\textwidth}
    \centering
    \includegraphics[width=\textwidth]{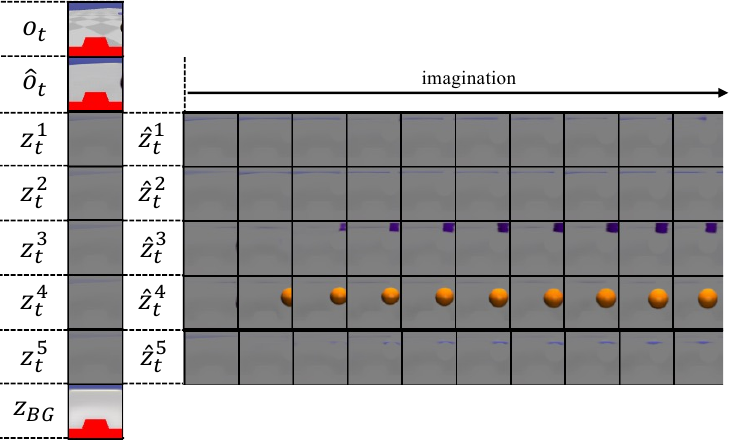}
    \vspace{-8mm}
    \caption*{(h) CarButton2}
    \vspace{5mm}
  \end{minipage}
  \begin{minipage}[t]{0.48\textwidth}
    \centering
    \includegraphics[width=\textwidth]{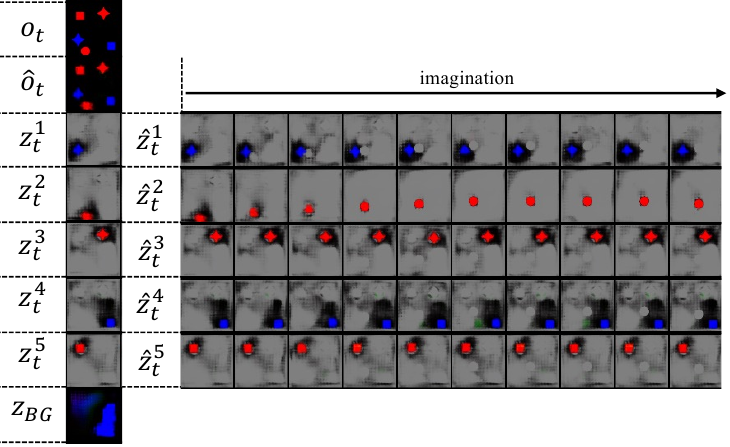}
    \vspace{-8mm}
    \caption*{(i) Object Goal}
    \vspace{5mm}
  \end{minipage}\hfill
  \begin{minipage}[t]{0.48\textwidth}
    \centering
    \includegraphics[width=\textwidth]{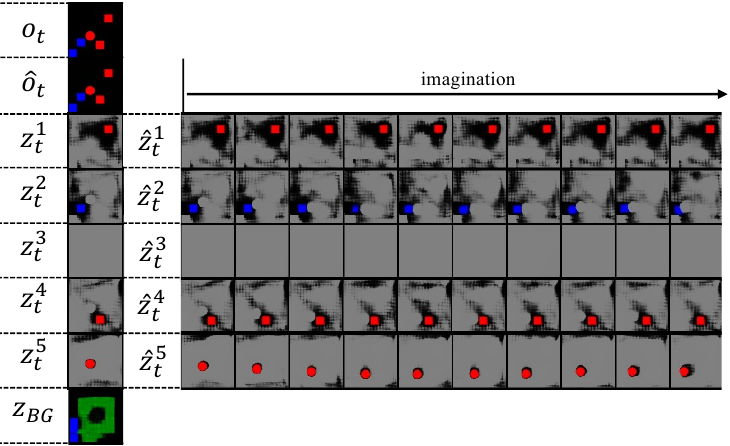}
    \vspace{-8mm}
    \caption*{(j) Object Interaction}
  \end{minipage}
  \begin{minipage}[t]{0.48\textwidth}
    \centering
    \includegraphics[width=\textwidth]{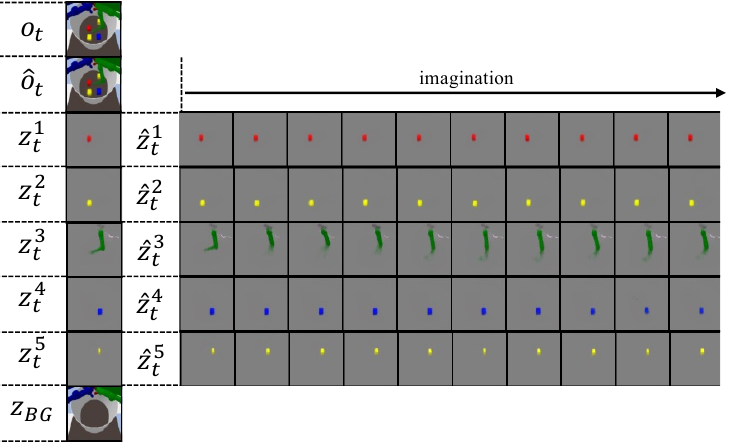}
    \vspace{-8mm}
    \caption*{(k) Object Reaching}
  \end{minipage}\hfill
  \caption{Examples of generated trajectories by the object-centric world model $\psi$ of STICA.
  (g)--(h) The Safety Gym benchmark.
  (i)--(k) The OCVRL benchmark.}
\end{figure}

\end{document}